\documentclass[sigconf]{acmart}

\AtBeginDocument{%
  }

\setcopyright{none}
\acmISBN{}
\usepackage{microtype}

\newcommand\cmu{$^2$}
\newcommand\penn{$^1$}
\newcommand\fordham{$^3$}

\title{Creation and Analysis of an International Corpus of Privacy Laws}

\author{
 \textbf{Sonu Gupta}\penn  \enspace
 \textbf{Ellen Poplavska}\penn  \enspace
 \textbf{Nora O'Toole}\penn  \\
 \textbf{Siddhant Arora}\cmu \enspace
 \textbf{Thomas Norton}\fordham  \enspace
 \textbf{Norman Sadeh}\cmu \enspace
 \textbf{Shomir Wilson}\penn \\
 \penn Penn State University, University Park, PA \\
 \cmu Carnegie Mellon University,  Pittsburgh, PA\\
  \fordham Fordham Law School, New York, NY \\
 {\tt \{sjg6104, esp5206, neo5060, shomir\}@psu.edu}\\
  {\tt  siddhana@andrew.cmu.edu} , {\tt tnorton1@law.fordham.edu} , {\tt  sadeh@cs.cmu.edu}
}

\renewcommand\footnotetextcopyrightpermission[1]{}
\renewcommand{\shortauthors}{Gupta et al.}
\date{}

\begin{document}

\begin{abstract}
The landscape of privacy laws and regulations around the world is complex and ever-changing. National and super-national laws, agreements, decrees, and other government-issued rules form a patchwork that companies must follow to operate internationally. To examine the status and evolution of this patchwork, we introduce the Government Privacy Instructions Corpus, or GPI Corpus, of 1,043 privacy laws, regulations, and guidelines, covering 182 jurisdictions. This corpus enables a large-scale quantitative and qualitative examination of legal foci on privacy. We examine the temporal distribution of when GPIs were created and illustrate the dramatic increase in privacy legislation over the past 50 years, although a finer-grained examination reveals that the rate of increase varies depending on the personal data types that GPIs address. Our exploration also demonstrates that most privacy laws respectively address relatively few personal data types, showing that comprehensive privacy legislation remains rare. Additionally, topic modeling results show the prevalence of common themes in GPIs, such as finance, healthcare, and telecommunications. Finally, we release the corpus to the research community to promote further study.\footnote{\url{https://usableprivacy.org/data}}
\end{abstract}





\maketitle
\pagestyle{plain}


\section{Introduction}

Privacy is a growing topic of attention for legislative and regulatory bodies around the world, and a growing number of documents produced by governments provide instructions for this topic. These government-issued instructions include legally binding documents such as laws and regulations, and also non-legally binding documents such as guidelines for following a law. Legal jurisdictions around the world have their own sets of government privacy instructions (GPIs, government privacy instructions), shaping the legal framework surrounding privacy within their particular jurisdictions.

At the same time, 
text analysis techniques have made it possible to study legal texts on a large scale. Prior efforts have studied legal text about privacy in the form of privacy policies, yielding insights for legal scholars and language models for the creation of privacy-enhancing technologies~\cite{hosseini2021analyzing, wilson2018analyzing, ravichander2021breaking}. Other efforts have applied NLP to legal text more generally~\cite{robaldo2019introduction, moreno-schneider-etal-2020-orchestrating}. However, despite growing interest in privacy, privacy law and NLP researchers have lacked a large-scale collection of texts of privacy documents from around the world. This stems from the non-trivial nature of this process. Often there are several official and unofficial versions of a document on the web. Among the official ones, governments also publish instructions to ``simply'' the adaptation of these laws,\footnote{\url{https://www.oaic.gov.au/__data/assets/pdf_file/0012/8013/privacy-safeguard-combined-chapters.pdf}}\footnote{\url{https://www.priv.gc.ca/en/privacy-topics/technology/online-privacy-tracking-cookies/tracking-and-ads/gl_ba_1112/}} which makes it challenging to distinguish them and legally enforceable documents. The task is further exacerbated by the absence of official translations of these laws. 

We address these challenges and present the Government Privacy Instructions Corpus (\textit{GPI Corpus}).
To the best of our knowledge, the GPI Corpus is the most comprehensive corpus of government privacy instructions to date, with natural language text in original languages and English.\footnote{In this study we focus on GPIs originally in English or translated to English, to match the authors' expertise. However, we acknowledge the importance of multilingual analysis, which motivated our inclusion of original non-English documents in the corpus for future use by ourselves or others.} The texts are paired with extensive metadata on the documents' electronic sources (i.e., URLs), relevant jurisdictions, dates of enactment, relation to international agreements, and other significant information. We coin the term \textit{government privacy instructions}, or \textit{GPIs} to characterize these documents, as the corpus encompasses laws, regulations, and government-issues guidelines and recommendations intended to instruct citizens, organizations, law enforcement, or lawmakers on actions to protect digital privacy. We include legally binding documents such as laws and government-produced non-legally binding documents such as guidelines in the corpus. Together, they provide a comprehensive view of the privacy instruction information provided by governments. In order to trace the history of such documents and the ways in which they may inherit vocabulary, concepts, and precedents from one another, the included documents comprise ones that are binding or relevant today, along with ones that have been in effect in the past. We also present the first large-scale study of GPIs using natural language processing (NLP) tools.
We examine temporal and topical trends in GPIs, showing a dramatic increase in attention to privacy over the past 50 years, a varied and nuanced distribution of mentions to personal information types, and a set of common themes that GPIs address.


We structure the rest of the paper as follows. In Related Work, we describe some prior efforts toward privacy text corpora and law corpora. In Corpus Creation, we describe the types of documents that comprise the GPI Corpus and the criteria, and how we gathered them from the web. In Distribution of GPIs, we show the differences in document availability and quantities of documents across geographic regions and across time. In addition, we use the metadata collected about the documents to make observations about the prevalence of GPIs over time and their availability in English. In Text Analysis, we study the distribution of mentions to personal data types across the corpus. We also apply LDA-based topic modeling techniques to extract the privacy topics discussed in the corpus. In the Discussion, we share the challenges of text analysis, and the limitations of this work. We conclude with Future Directions and Conclusions.


\section{Related Work}

We describe prior efforts toward language resource creation and NLP applications on four related domains of text: laws in general, legal documents, privacy policies, and privacy laws.

\textbf{Law Corpora:} Prior work has created international law corpora with varying foci. \citet{elliott} curated a master list of 779 international human rights instruments from 1863 to 2003 to highlight significant violations of those rights. \citet{labourRegulation} created a dataset of 63 labor laws to generate the Centre for Business Research - Labor Regulation Index (CBR-LRI) dataset. \citet{deakinsarkar} used this data to estimate the impact of labor regulation on unemployment. The authors further expanded this dataset to include 117 countries \cite{adamssimon}. Similar efforts have created datasets of non-English policies. \citet{impol} developed ImPol, a database, to estimate immigration policies in three European countries (France, Italy, and Spain) from 1960 to 2008. 

With the recent progress in NLP, language models have been created to apply law corpora to practical problems. Researchers used statutes of the US Internal Revenue Code to extract a set of rules along with a collection of natural language questions that can be answered correctly only by consulting these rules \cite{nils}. The authors also developed a StAtutory Reasoning Assessment dataset (SARA) for question answering and statutory reasoning in tax Law entailment. \citet{lame} proposed an NLP-based technique to extract concepts and relations from 57 French codes gathered from government websites that constitute 59,000 articles.

\textbf{Legal Document Corpora:} The analysis and interpretation of text dominates the field of law. Lawyers, judges, and regulators continuously compose legal documents such as memos, contracts, patents, and judicial decisions. Accordingly, there is a body of research about creating corpora of such legal documents. These corpora facilitate building Natural Language Understanding (NLU) technologies to assist legal practitioners. 

\citet{malik-etal-2021-ildc} introduce a corpus of Indian legal documents toward building an automated system for predicting the outcome of a legal trial as well as explaining the outcome. These automated systems can assist judges and help expedite the judicial process. Another similar study \cite{kalamkar2022corpus} annotates Indian legal documents for rhetorical roles, which has applications for both legal judgment prediction and legal summarization. \citet{chalkidis2021lexglue} create a benchmark dataset for various legal NLU tasks and evaluate different pretrained Language models on these tasks. There have also been similar efforts to develop legal documents corpora for non-English languages, such as \citet{mauri2021cadlaws}, who created a corpus of Canadian legal documents with legally equivalent texts in English and French, respectively. 

These corpora have enabled the creation of automated methods to interpret these legal documents. \citet{josi2022preparing} aims at automatic extracting text from signed PDF legal documents. Similarly, there has also been an interest in performing named entity recognition in legal domains \cite{puais2021towards}. Another work \cite{mistica2020information} aims at automating the extraction of information from legal judgments to assist lawyers on the case at hand. There has also been work in summarising legal text like court judgment documents to help legal professionals and ordinary citizens to get relevant information with little effort \cite{nguyen2021robust,jain2021summarization}. Similar efforts \cite{taylor2021towards,DBLP:journals/ail/YamadaTT19,xiao2021lawformer,avram2021pyeurovoc} have also been made to interpret legal documents in non-English languages.

\textbf{Privacy Policy Corpora:} Over the last decade, there has been significant growth in research about online privacy policies. The existence of data and high-quality annotations are essential for the application of both natural language processing and crowd-sourcing techniques to address the challenges posed by online privacy policies. This requirement has generated two threads in online privacy policy research: (i) annotation of privacy policy documents to facilitate future analysis and (ii) large-scale collection and analysis of privacy policies. The initial annotation attempts involved manual annotation of privacy policies by legal experts and crowd workers. Two such corpora are OPP-115 \cite{wilson2016creation} and APP-350 \cite{zimmeck2019maps}. OPP-115 consists of 115 web privacy policies (267K words) with 23K fine-grained data practices annotations. Although these corpora are relatively small, their annotations enable several researchers use them to train machine learning models to extract salient details from privacy policies \cite{sathyendra2017identifying,shvartzshanider2018recipe}.

In an attempt to create a larger corpus, \citet{harkous2018polisis} collected 130K mobile applications' privacy policies from the Google Play Store. 
Authors used the corpus to train a privacy policy-centric language model and built a set of neural network-based classifiers for both high-level and fine-grained aspects of privacy practices. In a similar effort, \citet{mukund} collected 1.4M privacy policies and developed a privacy policy search engine, PrivaSeer, which enables text query-based search across the collection. In follow-up work, authors \cite{srinath-etal-2021-privacy} trained a transformer-based language model using this corpus, resulting in the state of the art performance on classification and question answering tasks \cite{ravichander2019question, amos2021privacy}.

\textbf{Privacy Law Corpora:} In 2011, Graham Greenleaf performed the first global survey of data privacy laws and identified 76 countries that meet minimum international data privacy standards of international data protection and privacy agreements \cite{green2011}. After a decade, the seventh edition of this work \cite{greentable} expanded the global table to 145 and 23 countries with Data Privacy Laws and bills, respectively. This corpus has been used to analyze the momentum toward global ubiquity of data privacy laws \cite{green21analysis1}, the networks of data privacy authorities \cite{green21analysis2}, and progress for international data privacy standards \cite{green21analysis3}. World Legal Information Institute (WorldLII) developed a privacy research library that consists of links to case laws, commentaries, legislation, and more that several Legal Information Institutes originally maintain (LIIs) \cite{worldii}. DLA Piper, a global law firm, presents an overview of data protection laws for 89 jurisdictions \cite{piper}. 

Along with global corpora, there are studies of laws of specific regions. \citet{karldel} compared the data protection laws of Canada (PIPEDA), California (CCPA), the European Union (GDPR), and Quebec (the Quebec Private Sector Act). In \cite{african2016}, researchers analyzed the rising data protection systems in Africa concerning cultural differences across countries in addition to their socio-economic and political landscape. Authors compared 32 African data privacy laws at a fine-grained level against 30 features of data privacy law such as data quality, access, and collection \cite{african2020}. Further, researchers \cite{botha} highlighted the similarities and differences between the South African Protection of Personal Information Act (PoPI) and the international data protection laws. Similarly, in \cite{greenleaf2014asian}, the author discussed and analyzed Asian data privacy laws in-depth. 

Our work closely aligns with the previous work by Greenleaf. We take a broader perspective of the data protection laws and broaden the inclusion criteria 
to extend our corpus by including more jurisdictions and documents (e.g., guidelines). In addition, all the above efforts present only qualitative analysis. In contrast, we employ both quantitative and qualitative methodologies. We also leverage NLP tools and machine learning algorithms to study this large-scale corpus. Lastly, unlike previous work that shared the list of the names of these documents, we share the original text of all the documents. We also consider the multi-lingual dimension and share both the non-English and English translations.


\section{Corpus Creation}

\begin{figure*}[t]
	\centering
	\captionsetup{justification=centering}
	\includegraphics[width=0.8\textwidth]{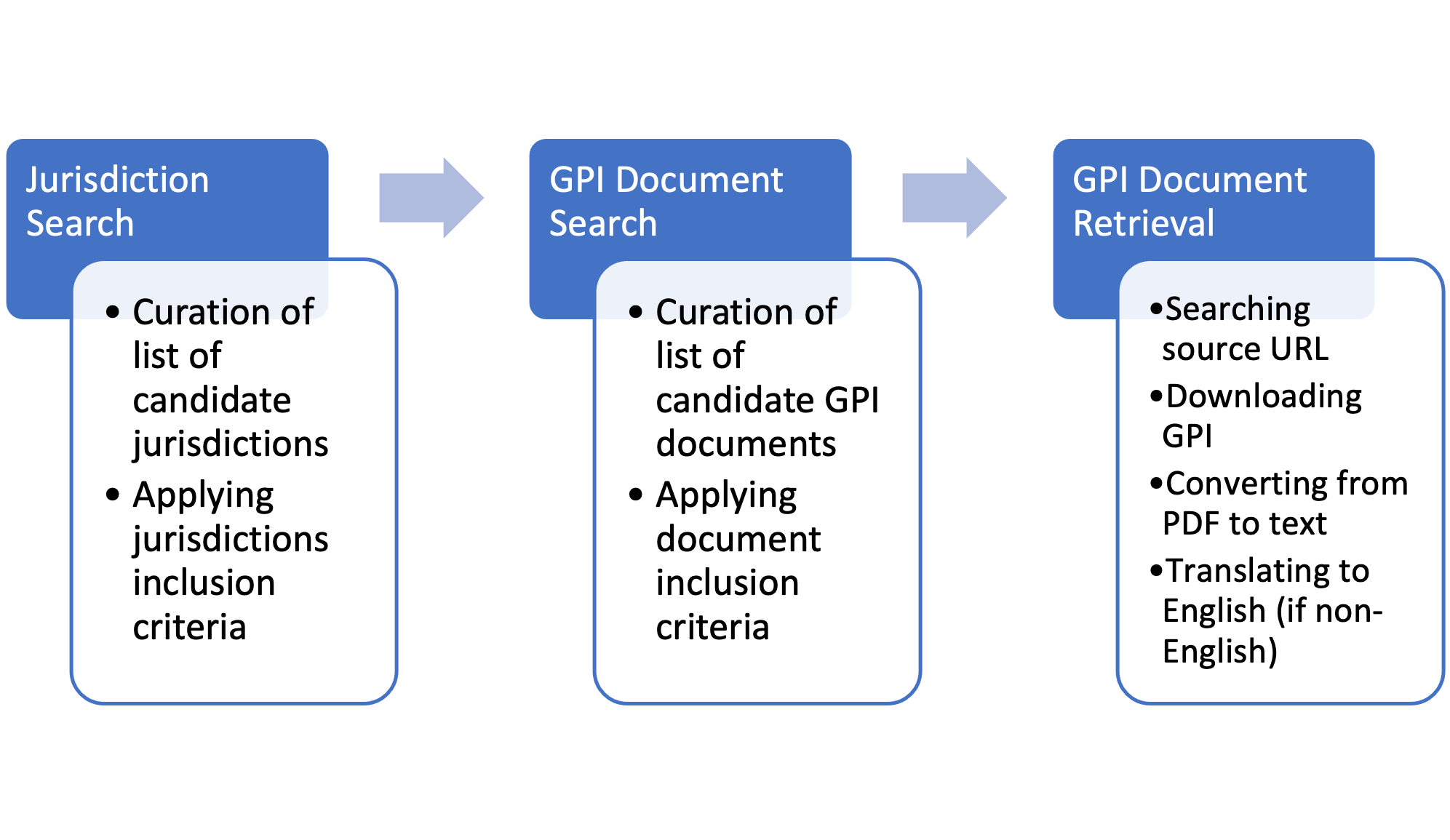}
	\vspace{-1cm}
	\caption{The end-to-end pipeline of the creation of GPI corpus.}
	
	\label{fig:flowdiagram} 
\end{figure*}

Corpus creation required a series of overarching tasks: searching by jurisdiction for document that ought to be included in the corpus, determining precise jurisdiction and document inclusion criteria, manually collecting GPIs for the selected jurisdictions from the internet, and categorizing these documents into three subdivisions. We summarize the entire pipeline of the corpus creation tasks in Figure \ref{fig:flowdiagram}.

\subsection{Jurisdictions} \label{sec:jur}

Intending to achieve extensive coverage of nation-level jurisdictions worldwide, we curate a list of candidate jurisdictions prior to collecting their GPI documents. First, to build this list, we defer to the existing work by \citet{greenleaf2019global}, and leading legal experts that provide such information with their online legal resources such as Data Guidance \cite{data_guidance} and DLA Piper \cite{dlapiper}.
The jurisdictions mentioned in these sources serve as the starting point for our work, and for the duration of the corpus creation process, we often refer to them. For the sake of simplicity, we call them \textit{reasoning documents}. Next, we instate a series of inclusion criteria 
to scope our list of jurisdictions.

The initial list of jurisdictions derived from Greenleaf's table is limited. We believe privacy researchers can benefit from a more comprehensive list with better coverage of documents from around the world at a national level. This requires the development of a set of rules to filter out the jurisdictions that are outside the scope of our work. These criteria facilitate our manual search for the jurisdictions across the web to expand our initial list, weighing each jurisdiction against these criteria to decide whether it ought to be included to make the corpus representative of a consistent group of jurisdictions that fulfill certain requirements.

To develop a criterion to represent all available documents from various nations, as well as all categories of non-nation locales represented in the \citet{greenleaf2019global}, we decided to focus on country-level jurisdictions together with a few special categories. We include a jurisdiction if it satisfies one of the two requirements: (i) it is a country recognized as either a member or observer state of the United Nations by at least one other member state as of 2020, and (ii) a jurisdiction falls into the following special categories: (a) self-governing British Overseas Territories (Bermuda, Gibraltar, Cayman Islands), (b) crown dependencies (Guernsey, Jersy, Isle of Man), (c) Chinese Special Economic Regions (Macau and Hong Kong), (d) Qatar economic free zones (Qatar Financial Centre), (e) United Arab Emirates economic free zones (Abu Dhabi Global Market, Dubai International Financial Centre, Dubai Healthcare City), and (f) the states which are not recognized as UN members or observers (The Republic of China (Taiwan) and Kosovo). We also include one US state (California), with a rationale explained below.

A subset of the jurisdictions from the special jurisdictional categories is reverse-engineered from the list of jurisdictions originally mentioned in the Data Privacy Tables developed by \citet{greenleaf2017global, greenleaf2019global}.  
To ensure consistent presentation of all the jurisdictions from each of the described jurisdiction types, we add several individual jurisdictions 
that are not part of Greenleaf's table. We further expand our list of relevant documents with the help of information present in the documents that satisfy our inclusion criteria. With an exception, we include one US state, California, in our corpus, due to its significance and weight in defining privacy legislation that impacts the entire US economic system \cite{the-national-impact-of-ccpa}. 


In summary, this process resulted in a list of 182 jurisdictions, with 166 at the country level (86\% of 193 United Nations member states \cite{memberstates}).
 For the remaining 27 countries in United Nations member states, 
either no GPI exists or it was irretrievable on the internet.

\subsection{Government Privacy Instruction Documents} \label{sec:gipdocs}

To create an initial list of candidate documents, we refer to the list of 132 
privacy laws collected by Greenleaf \cite{greenleaf2019global} and, by default, include documents present in it. Several other documents are included because of their inclusion in online resources compiled by legal experts for public viewing and use, pertaining to the applicable privacy legislation and guidelines within each of several jurisdictions. Iteratively, we extend this list if an existing document of this list points to other candidate documents. 
Upon discovery of additional documents, we apply our GPI document inclusion criteria
to determine whether they should be included in the corpus.

Our goal is to curate an initial exhaustive list of all candidate GPI documents. 
However, we need to filter out the documents that fall beyond our scope. We develop two sets of rules based on document type and source to address this. All the documents to be included must satisfy at least one criteria from both sets of the rules.

\textbf{Criteria based on document type ---} 
Each document must meet at least one of the following inclusion criteria:
\begin{enumerate}
\item The document is legally enforceable (or once-enforceable and now defunct, or assumed to be enforceable upon some future date of effect), which is promulgated in a complete state to the general public for the purposes of awareness of the law and enforcement if it is in force, which may include laws and regulations.
\item The document contains rules, clarification, or similar resources directed towards lawmakers or law enforcement for the purposes of enforcing the aforementioned document.
\item The document contains a non-enforceable list of guidelines, which serve as official guidance directed towards the general public, or specific sectors of the public, for the purposes of advising them on how to comply with a document of another type.
\end{enumerate}

\textbf{A detailed description of excluded document types ---} These document types exclude case law, which establishes legal precedents through individual court decisions. Although such cases are valuable pieces of information and form precedents for decisions regarding compliance with laws related to privacy, they neither form an explicit legal directive or instruction nor a document explicitly instructing the reader about how to enforce or comply with such instructions. Additionally, due to the overwhelming scope and limited resources for acquiring case law notices or summaries globally, case law is categorically excluded from this work.

This corpus also excludes discussions of legal rationale unaccompanied by content that matches the aforementioned document types. Much like case law, discussions and arguments explaining the rationale behind a legal directive are a malleable resource that can be used to understand the application of the law. However, we exclude them because such documents also do not provide any direct instruction or guidance to the reader and, instead summarize lawmakers' theoretical decisions.

The final notable type of document excluded from this work is national constitutions, which provide established principles with significance both in their own right as legal documents and as a potent precedent for other laws developed in the country. We categorically exclude national constitutions because, in the overwhelming majority of cases, allusions to a right to privacy in a constitutional document were found to lack actionable details regarding expectations, instructions, or enforcement. Thus, although such mentions within national constitutions may act as a guiding principle in the development of subsequent legal documents regarding privacy, we find that these constitutional documents do not provide enough instruction to lawmakers, enforcers, or citizens regarding privacy to be a meaningful and effective part for our corpus. 

\textbf{Criteria based on source type ---}
The documents must meet any one of the following inclusion criteria for document source:

\begin{enumerate}
\item The document contains more content than a notification containing some update regarding the legal status of another document. A decree that says only that a different law is now in effect, providing no further guidance or substance, is excluded. 

\item The document is released by a government entity, such as (but not limited to) an executive order released by a president, a law passed by a congress or parliament, or a set of rules released by a government agency. Documents released by non-government entities, such as rules released by corporations and non-profit organizations for the internal governance of data privacy, guidelines released for the general public, and others, are excluded.

\item The document is released to the general public with the intent of circulating the document in its current, complete state for the purposes of understanding or enforcement of the document. Such circulation resources may include government websites and legal journals. 
This implies that the following types of documents are excluded.

\begin{enumerate}
    \item Private documents are not meant for such release to the public.
    
    \item Rules that describe internal procedures not directly relevant to the privacy laws and concepts in question, such as documents that merely describe which agencies or positions are charged with particular enforcement duties, are not included in this corpus. This is because these documents do not provide meaningful context into how the meaning of law itself is interpreted and enforced.
    
    \item Activity reports of government agencies, meant primarily for internal review and as a resource regarding the state of enforcement. Because of their conceptual removal from the types of documents of interest to the researchers.
    
    \item Strategy and action plans designed for internal use by enforcement agencies. 
    
    \item Enforcement decisions and records of fines. This is because they are notices aimed towards the specific audience of a given punished entity, without a desired audience of the general civic public.
\end{enumerate}

\item While future versions of the document may be released with changes, the document is released within its given form with the understanding that this form is immutable and is to be understood as-is until further documents are released to update it. This implies that the following types of documents are excluded.
\begin{enumerate}
    \item Bills and similarly unfinished documents released in various drafts for the purposes of transient public forum discussion.
    \item Forms, software tools, and other tools that require active constituent participation for effective use. As the form of these artifacts extends beyond the static, immutable document states that we wish to analyze here.
\end{enumerate}

\item The documents are promulgated in their included region by or before December 31, 2020. We set that date significantly in the past to promote higher recall in the final years of the corpus, recognizing that documents from some jurisdictions are not immediately available online.
\end{enumerate}


\subsection{Document Collection} {\label{sec:col_doc}}

If a document is deemed fit to be included within the corpus, we look for the source of the downloadable version on the web. We begin the search from our set of reasoning documents. We are able to find a few direct sources to downloadable documents mentioned in the reasoning documents. However, in most cases, they do not provide direct links to the source documents we sought to include; for several documents, only the associated legal document titles are mentioned without a link to the law or other legal document. This is especially true for documents that are not originally composed in the English language, in which case we seek to collect both an original language version and, if available, a human-translated English version of the document. Since few reasoning documents are linked to only one language version of a document or to no version, we leverage the mentioned legal document titles to search for the sources of other legal documents that are not present in the reasoning documents. 

We collect all the documents within this corpus manually from the internet using the document inclusion criteria described above. The collection activities include locating, downloading, and uploading documents to the repository and recording the metadata. It took two researchers from our team approximately 60 
labor hours combined to complete these tasks. Since this process of manual collection is conducted within the broad scope of any candidate documents we might find on the internet, rather than a closed list of automatically retrieved results, we take every caution to apply the inclusion criteria described above comprehensively to all the documents we find in the process of collection.

We download all the relevant documents from the web in PDF, to preserve visual formatting. This includes both the document in the original language and their English translation wherever available. Since we cannot directly extract the text content from PDF documents, we convert them into a text file format (.txt) using Apache Tika \cite{tika}. 
We attempt to use OCR \cite{mori1999optical} to convert scanned documents, although the software was not always successful. In instances when we were only able to collect a non-English version of a document, we translated its text file to English using online tools. We elaborate on the issues and challenges of the translation process in \S \ref{ref:subsec_trans}.

\subsection{Subdivisions of the Corpus} \label{sec:subdiv}

We divide the GPI Corpus into three sets: the primary set, the untranslatable set, and the irretrievable set. This categorization is necessary to segregate the documents based on their availability for content analysis. The primary set comprises 
documents within the canonical body of the corpus. It includes documents that exist as an available English language text within the corpus, whether this text is the original document, a human translation, or a machine translation. Next, as the name suggests, the untranslatable set contains documents for which we present original, non-English-language text, and we are unable to translate into a usable English-language version. Lastly, the irretrievable set is a list of documents we sought to include in the corpus but are ultimately unable to retrieve from the internet in a usable state either in English or in some original non-English language. The failure modes for the untranslatable set and the irretrievable set are detailed in the \S \ref{ref:subsec_trans} and \S \ref{ref:subsec_removed_docs}. In addition, if available, we retrieve the metadata (e.g., the year of enactment, if in effect or repealed) for all the documents present in these three sets and use it for the temporal analysis of the corpus.


\section{Distribution of GPI\texorpdfstring{\MakeLowercase s}{s}}

The process described in the above section results in 1,043 documents for analysis. Based on the coverage, we classify the jurisdictions into three categories, (1) National, (2) International, and (3) State/Province. As shown in Table \ref{tab:tb_diversity}, the majority of the documents cover national level jurisdictions and contribute to 95.49\% of the documents in the corpus with 166 unique jurisdictions, whereas 161 distinct countries make up 91.27\% of the documents. Countries that participated in various international agreements also have their own unique sets of documents within the jurisdiction of their own country.

The collected documents are laws and regulations, rules, guidelines, and other government-released documents, communiques, notices, circulars, orders, decrees, and decisions. Each document is in its current state or the latest state of revision if any. The latest revision date is recorded for each document from the law category. We add the promulgation date as the last revision date if no revisions have occurred.

\begin{table*}[t]
\centering
\captionsetup{justification=centering}
\caption{Summary of corpus composition. In the Coverage Type column, N, I, and S/P represent National, International, and State/Province jurisdictions, respectively. }
\begin{tabular}{lllll} 
\toprule
Jurisdiction Type & Coverage Type &\# Unique Jurisdictions & \# Documents & Examples\\
\midrule
Countries & N & 161 & 952  & Albania  \\ 
British Overseas Territories & N & 3 & 24 & Cayman Islands \\ 
Crown Dependencies & N & 2 & 20  & Isle of Man \\ 
International Organizations & I & 3 & 13  & United Nations \\ 
Privacy Frameworks  & I & 3 & 3 & Asia-Pacific Economic Cooperation\\ 
Intergovernmental Organizations & I & 3 & 4 & US + 23  Countries \\
Special Economic Zones & S/P & 4 & 8  & Qatar Financial Centre \\ 
Special Administrative Regions & S/P & 2 & 16 & Macau \\
State  & S/P & 1 & 3 & California (USA) \\ 
\bottomrule
\end{tabular}
\label{tab:tb_diversity}
\end{table*}

\begin{figure*}[t]
	\centering
	\captionsetup{justification=centering}
	\includegraphics[width=0.6\textwidth]{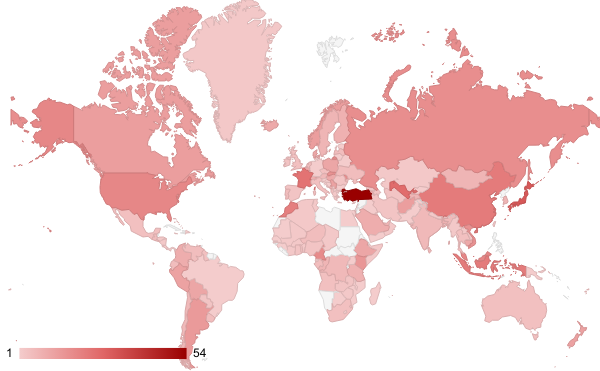}
	\caption{World map representing the number of enforceable privacy laws from each country in GPI corpus.}
	\label{fig:map} 
\end{figure*}

We show the number of enforceable privacy laws in our corpus per country as a map in Figure \ref{fig:map}. We observe that Turkey has the most GPIs, followed by Japan, Uzbekistan, and France. We explore the reason for Turkey's exceptional number in Section \ref{ref:temp_dis}. The map also shows several countries that have zero GPIs in the corpus, particularly in Africa.


\subsection{Translations} \label{ref:subsec_trans}

The corpus comprises documents in 54 languages, with 37.13\% documents in English, making it the most common language. In addition, 85 documents are written in both English and the native language of the region in a single document. Chad is the only exception where the document is written in two languages (French/Arabic), and neither of them is English. There are six languages that only appear in combination with the English language. For instance, all the 11 documents from Malta are written in Maltese/English.

We attempt to create English translations for all the non-English documents in the corpus, to establish a uniform natural language for text analysis. Based on the translation, we divide the corpus into four classes: (1) Originally in English, (2) Official translation, (3) Unofficial translation, and as the name suggests, and (4) Google translation. 
If a document is in a language other than English, we seek an official English translation provided by the source of the official non-English document. Sometimes, official sources provide a translated version but call it an unofficial document for legal purposes.
In the absence of the availability of such translations, we turn to international privacy expert sites, with exact sources noted in the corpus metadata. However, if the translation is still unavailable, we use translation tools like Google Translate \cite{go_trans}. 
As we present in Table \ref{tab:trans}, 56.18\% documents within the corpus do not contain a translation. It emphasize the difficulty in searching and accessing the non-English GPIs due to language barrier.

\begin{table} 
\centering
\captionsetup{justification=centering}
\caption{Distribution of the sources of English
translations.}
\begin{tabular}{lll} 
\toprule
Translation Type & No. of Documents & \% of Total \\ 
\midrule
Originally English & 372  & 35.67\%\\ 
Official    & 66  & 6.33\%\\ 
Unofficial & 19  & 1.82\%\\ 
Google Translated & 586  & 56.18\%\\ 
\bottomrule
\end{tabular}
\label{tab:trans}
\end{table}

Out of 671 non-English documents, only 41 have English titles available in the original documents. We utilize these English titles to locate the source of the English version of the document on the web. In the absence of non-English titles, we turn to Google translate. However, it fails to provide a usable translation for a few titles in the Russian language. Therefore, we turn to Yandex Translate \cite{yandex_trans}. Yandex 
is a Russian technology company that provides internet-related products and services \cite{yandex}. We also utilize Yandex Translate to scan the contents of a few non-English websites and look for the required document. 

\subsection{Temporal Distribution}  \label{ref:temp_dis}

\begin{figure*}[t]
	\centering
	\captionsetup{justification=centering}
	\includegraphics[width=0.8\textwidth]{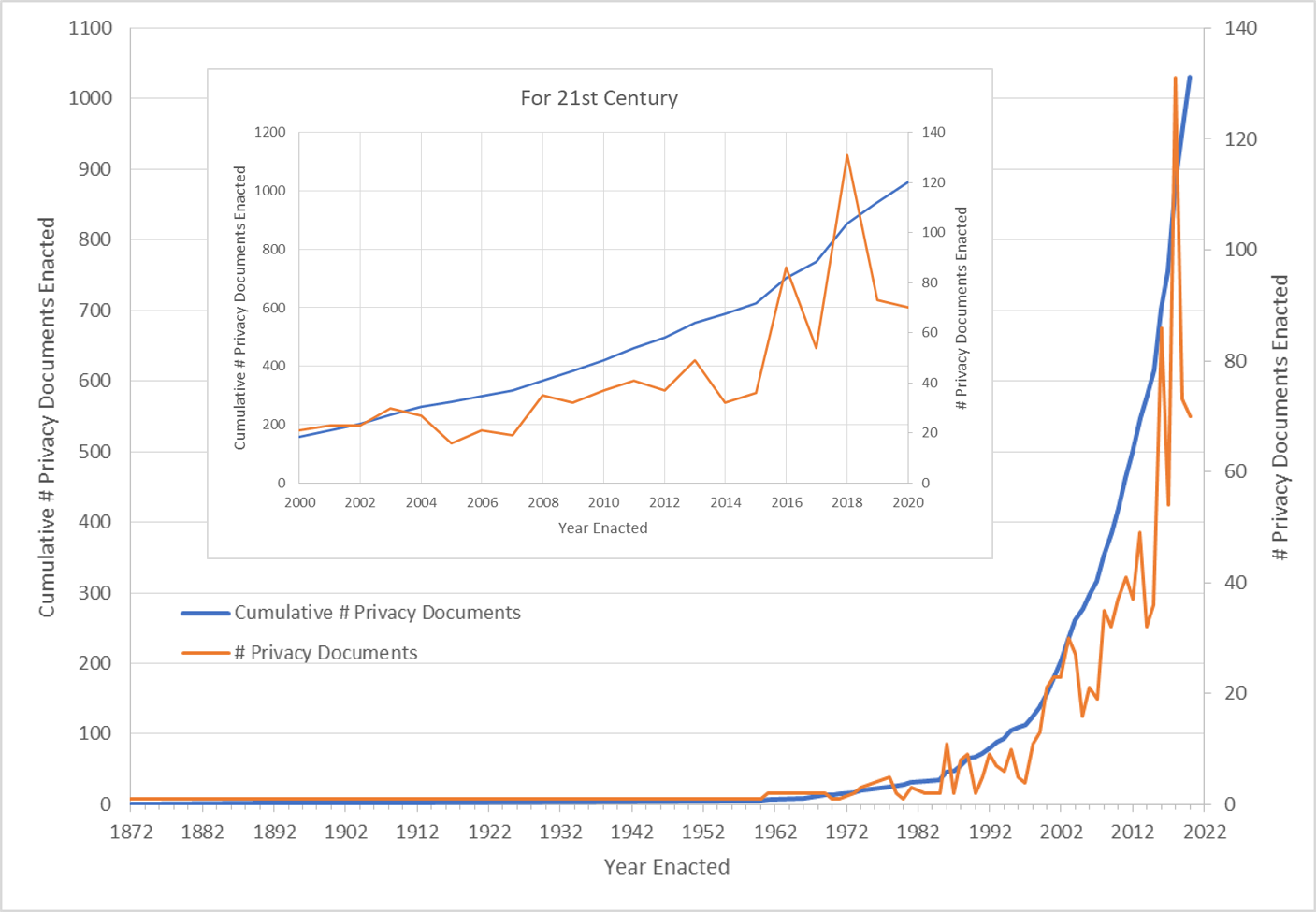}
	\caption{GPIs enacted over time.}
	\label{Figure3} 
\end{figure*}

We examine the distribution over time of the creation of GPIs, as the corpus contains documents dated as early as 1872 and as recently as December 2020. We illustrate the pace of GPIs enacted over this date range in Figure \ref{Figure3}. It is a  dual-vertical axis graph where the left and right vertical axes show the cumulative and total number of GPIs enacted over the years, respectively. We perform a chi-square test for the goodness of fit and detect a trend, starting in 1966, that the pace of GPIs grows exponentially with rate parameter (\(\lambda\)) equal to 0.054. 
We also find a sharper exponential growth in the 21\textsuperscript{st} century with (\(\lambda\)) equal to 0.036. It should be noted that in a few documents first written in the late 80s and early 90s, the data privacy statements were included only after revisions. For example, the criminal code of Finland was enacted in 1889, but a data privacy section wasn't added to it until 2015. 

We notice two discernible peaks in 2016 and 2018. In 2016, 86 GPIs were issued, out of which 32 were published in Turkey only. After the failed July 15, 2016 coup attempt \cite{coup}
in Turkey, several emergency decrees were published \cite{bazy2016opinion}. We speculate that the coup attempt was the cause of the sudden increase in GPIs in Turkey. We also observe that the largest number of GPIs were issued in 2018, with a total of 131 GPIs in 67 distinct jurisdictions. We speculate that the enactment of GPDR in early 2018 may have caused the increase in the new documents as 16.79\% of documents explicitly mention GDPR in their title. Additionally, GDPR may have encouraged the presence of documents to be in digital format and available over the web. 
We note a continuous increase every decade, with the most number of jurisdictions receiving their first GPI between 2010 and 2019. There are 62 such jurisdictions, including countries like Costa Rica, Barbados. 

\subsection{Excluded Documents} \label{ref:subsec_removed_docs}

There were 153 documents that we attempted to include in the primary set but were ultimately unable to add 
due to several failure modes.
This led to the categorization of documents into three sets as discussed in \S \ref{sec:subdiv}.
We present this categorical distribution in Table \ref{tab:table-remov_doc}.
The primary set consists of 1,043 documents 
that are either originally in English or translated into English.
For 14
documents that could not be fully translated into English and the percentage of English words within the document was less than 95\%, 
we added them to the untranslatable set\footnote{We use langdetect \cite{langdetect}
to measure the English language content in a document.}, and it received only metadata analysis. For three documents, Google Translate could not interpret the entire document due to a document size constraint \cite{google_limit}. We split these documents into smaller chunks in such scenarios and then perform the translation. Further, Google Translate does not offer services for all languages in the corpus. One such language not offered by Google Translate is Dzongkha, a language used in one of Bhutan's GPIs in the corpus. Lastly, we mark a document irretrievable if we are unable to locate the original English or Non-English document on the web in a machine-readable format. This resulted in the removal of 139 documents from the corpus. Below are other failure modes that contributed to the irretrievable set:

\begin{enumerate}
    \item Errors with Optical Character Recognition: Running a scanned document through Optical Character Recognition (OCR) \cite{srihari2003optical} results in a machine-readable text data. We utilize OCR for documents originally present on the web in a scanned version. However, 58 documents could not be accurately or completely converted. 
    
    \item Document Not Found: This includes 50 documents mentioned in several reasoning documents or other GPIs, but we could not find them on the web. 
    
    \item Page Not Found: This includes 20 documents for which we are able to source a URL, but the available URL resulted in a 404 HTTP error.
    
    \item Suspicious URLs: If the web browser blocks access to a URL, we exclude it. We have two such links associated with the jurisdiction of Montenegro.
    
    \item Cannot Access: Nine documents are not available on the web directly and require a paid subscription of a service like Guidance Notes \cite{malta-national-gdpr-implementation}.
\end{enumerate}


\begin{table}[H]
\centering
\captionsetup{justification=centering}
\caption{Summary of the subdivision of the documents.}
  \label{tab:table-remov_doc}
  \begin{tabular}{lll}
    \toprule
    Set Name & No. of Documents & Percentage \\
    \midrule
    Primary Set & 1,043 & 87.21\% \\ 
    Untranslatable Set  & 14 & 1.17\%\\ 
    Irretrievable Set & 139 & 11.62\%\\ 
  \bottomrule
\end{tabular}
\end{table}


\section{Text Analysis}

We use text analysis methods to examine the composition of the GPI Corpus and trends in GPIs.
In the rest of this section, we use \textit{corpus} and \textit{primary set} interchangeably. 
Table \ref{tab:table-corpus-word-length} shows some summary statistics for the corpus. 
The documents vary widely in length, from 40 words to 509,094.


\begin{table}[H]
\centering
\captionsetup{justification=centering}
\caption{Word count summary statistics for the GPI Corpus.}
  \label{tab:table-corpus-word-length}
  \begin{tabular}{ll}
    \toprule
    Minimum words in a document & 40\\ 
    Median words per document & 6,657 \\ 
    Mean words per document & 16,112 \\ 
    Maximum words in a document & 590,094 \\ 
    Total words in corpus &  16.82M \\ 
  \bottomrule
\end{tabular}
\end{table}


To explore mentions to technologies in GPIs, we hand-curate a list of 32 keywords used to describe technologies relevant to consumer privacy.
We started with a small list of words referring to common technologies (e.g., \textit{computer}, \textit{website}) and after multiple iterations of discussion with privacy and legal experts, we expanded it to 32 keywords. 
For text analysis, we convert all the keywords into lower case and their singular forms (e.g., ``Emails'' is changed to email). We perform the same preprocessing steps on the corpus documents and then calculate the statistics.
We consider the text of primary set documents and find that out of 1,043 documents, 73 documents do not contain any of the keywords. We present top ten most frequent 
keywords in Table \ref{tab:spec_tech}.
We observe that \emph{email} and \emph{phone}, two of the oldest methods of electronic communication in our list, lead the frequency ranking with a large gap before the fourth (\emph{computer}). 

\begin{table}[h]
\centering
\captionsetup{justification=centering}
\caption{Top ten technologies most frequently mentioned in the corpus.}
  \label{tab:spec_tech}
  \begin{tabular}{ll}
    \toprule
    Keywords & Frequency\\
    \midrule
    Email & 958\\ 
    Phone & 896 \\ 
    Network & 683 \\ 
    Computer & 523\\ 
    Telephone & 394\\ 
    Electronic Communication & 323\\ 
    Website & 285\\ 
    Biometric & 252\\ 
    Internet & 222\\ 
    Television & 205 \\ 
  \bottomrule
\end{tabular}
\end{table}




\subsection{Personal Identifiable Information}

Personal identifiable information (PII) includes any information associated with an identified or identifiable living person in particular that can be connected to an identifier such as a name, national identification number, email address, and more \cite{pdatadef}.  
GPIs often explicitly include a descriptive definition of PII at the beginning of the document. For instance, \textit{``...personal information refers to personal information: (1) About an individual’s race, ethnic origin, marital status, age, color, and religious, philosophical...''} - Data Privacy Act (2012), Philippines \cite{dprivacy}.

We examine the distribution of mentions of PII types\footnote{In this paper, we assume a broadly inclusive position for what personal information counts as PII, as prior work has shown identity can be reconstructed from a variety of information types \cite{henriksen2016re}.} in GPIs and their trends over time to identify differences in attention and temporal trends. 
To do this, we create a list of 154 PII keywords with the help of following official sources:
\begin{itemize}
    \item The U.S. National Archives and Records Administration \cite{archi}
    \item The U.S. National Institute of Standards and Technology (NIST) \cite{nist}
    \item The U.S. Federal Trade Commission (FTC) \cite{ftc}
    \item The European Commission \cite{euc}
\end{itemize}

We expand the list by including several country-specific alternates for each PII keyword. For example, national ID schemes are known by many names, including \textit{Social Security Number} in the US\footnote{\url{https://www.ssa.gov}}, \textit{Documento Nacional de Identidad} in Argentina\footnote{\url{http://www.interior.gob.es/web/servicios-al-ciudadano/dni}}, and \textit{Aadhaar} in India\footnote{\url{https://uidai.gov.in}}.  
We also include variations for non-country specific terms. For example, a name could be mentioned as a middle name, first name, last name, mother’s maiden name, surname, and more such variations. 

We place our PII keywords into 15 unique categories, as shown in Table \ref{tab:pii_cats}, 
and create a miscellaneous category for four keywords that do not fit any of these themes. The complete list will be provided upon acceptance. To perform analysis, we convert all the keywords to lower case and their singular forms (e.g., ``Ages'' is changed to age). Similarly, we consider abbreviations of the keywords as well. For instance, we look for ``mobile number'' as well as ``mobile no.'' and count them as the same entity. We perform the same preprocessing steps on the corpus documents and then calculate the statistics.

\begin{table}
\centering
\captionsetup{justification=centering}
\caption{PII types along with its sample keywords.}
  \label{tab:pii_cats}
  \begin{tabular}{ll}
    \toprule
    PII Types & Example Keywords\\
    \midrule
Finance &  Credit Score, Bank Account Number\\ 
Work & Salary, Employment Information\\ 
Health & Medical History, Prescription\\ 
Biometric & Retina Scan, Fingerprint, Voice-print\\ 
Genetic  & DNA, DNA profile, genetic information\\ 
Bio./Demographic & Name, Age, Family Information\\ 
Race/Ethnicity  & Racial Origin, Ethnic Origin, Ethnicity\\ 
Beliefs & Religious, Philosophical, Political\\ 
Technology  & Computer Information, Text Message\\ 
Tracking IDs  & IP address, Cookie ID, \\ 
Govt./Personal IDs  & State ID, Passport, Driver's License\\ 
Location  & Geo-location, Home Address, Zip Code\\ 
Contact  & Mobile Number, Fax Number\\ 
Photo &  Photographic Image, Personal Photo\\ 
Misc. & Criminal History, Union Membership\\ 
  \bottomrule
\end{tabular}
\end{table}

\begin{figure}[h]
	\centering
	\captionsetup{justification=centering}
	\includegraphics[width=0.5\textwidth]{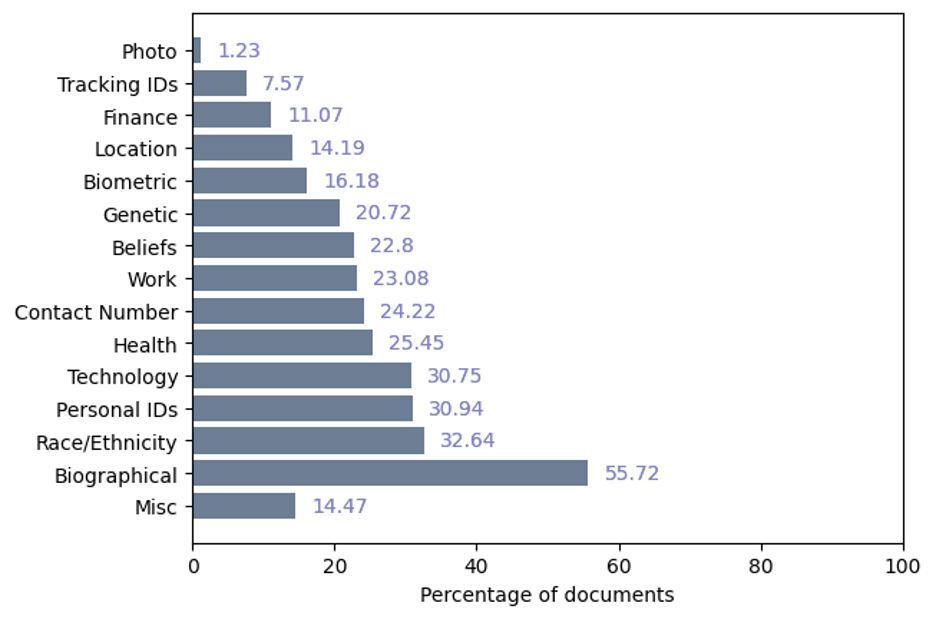}
	\caption{Distribution of the GPI documents per PII types.}
	\label{fig:pii_distribution_1} 
\end{figure}
\begin{figure*}
	\centering
	\captionsetup{justification=centering}
	\includegraphics[width=0.6\textwidth]{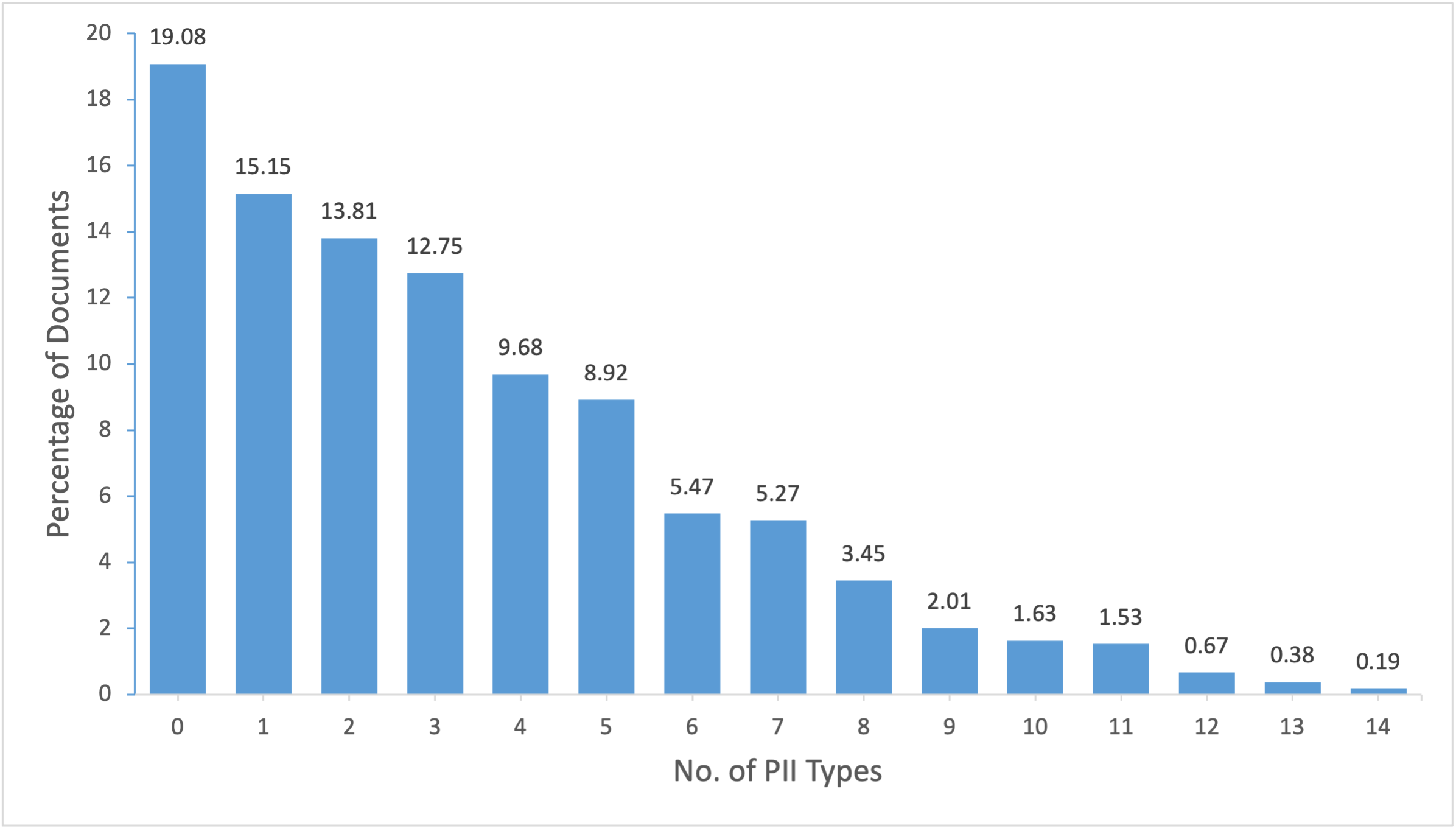}
	\caption{Distribution of the number of PII types present per document.}
	\label{fig:pii_distribution_2} 
\end{figure*}

Figure \ref{fig:pii_distribution_1} shows that for every category besides biographical, the percentage of documents that contain keywords from the category is less than half.
We also observe that the tracking IDs, a category that contains relatively technical terms compared to other categories, do not often appear in the GPIs. This suggests that laws refrain from committing to regulating specific representation of the information. Additionally, Figure \ref{fig:pii_distribution_2} represents the number of PII types present per document in the corpus,
shows that privacy legislation that addresses a large number of different PII types is rare. About 60.79\% GPIs represent three or fewer PII types. There are only 0.19\% GPIs that are comprehensive enough to cover the 14 PII types listed in the Table \ref{tab:pii_cats}. 
According to this measure, the two most comprehensive privacy laws are California Consumer Privacy Act (CCPA) and California Privacy Rights Act (CPRA).

\begin{figure*}[t]
	\centering
	\captionsetup{justification=centering}
	\includegraphics[width=0.6\textwidth]{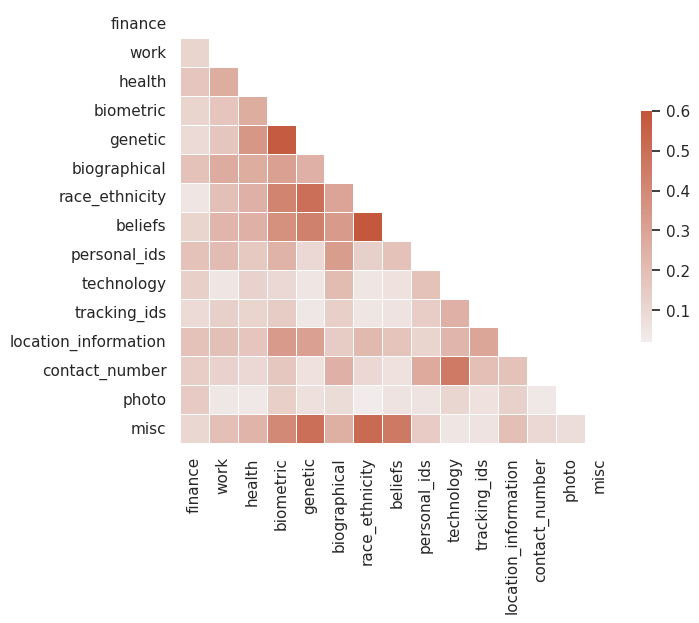}
	\caption{Pearson correlation between the occurrence of the PII types in the corpus.}
	\label{fig:pii_corr} 
\end{figure*}

We further compute the pair-wise correlations between the occurrence of PII types in each document to investigate the linear relationship between PII types. We show the results in Figure \ref{fig:pii_corr}. We note that the correlations between all the PII types are always positive but differ for all the PII pairs. Although the correlations are positive, they are below 0.6, showing that no one pairing is dominant. The highest correlation is between Race/Ethnicity and Beliefs.
GPIs that fall under this scenario include documents
from 95 distinct countries and cover 16.29\% of documents in the corpus. We also observed a high correlation between Genetic and Biometric PII types, which share a biological focus.

\begin{figure*}[t]
	\centering
	\captionsetup{justification=centering}
	\includegraphics[width=1\textwidth]{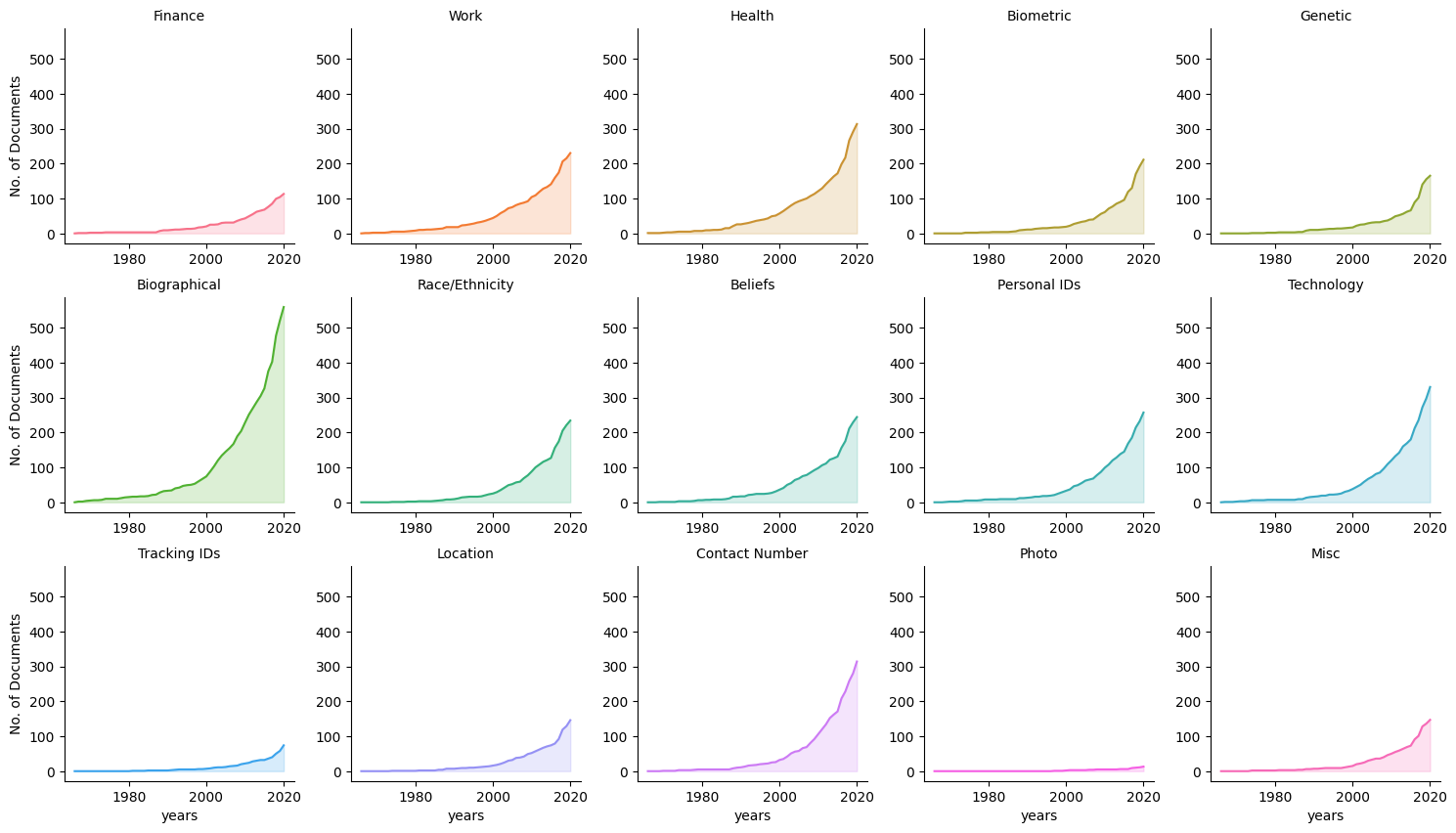}
	\caption{Trends in the mention of personal information keywords in the GPI corpus of 1,043 documents. The vertical axis represents the cumulative no. of documents that contain keywords for a PII type and the horizontal axis is the year in which a document is enacted.}
	\label{fig:pii_timeseries} 
\end{figure*}

In Figure \ref{fig:pii_timeseries} we show that all the PII types exhibit increase (i.e., the second derivative is positive) in frequency over time, but the rate of increase varies across the categories. For instance, we observe that for the ``Biographical/Demographic'' and ``Contact Number'' increase has been rapid; however, for ``Tracking IDs'' and ``Photo'' the rate of increase is more sedate.

\subsection{Topic Modeling}

To explore the range of the topics covered in the GPIs, we turn to algorithmic methods. In machine learning, topic modeling is an unsupervised learning technique that identifies major themes or topics in a collection of documents. We leverage Latent Dirichlet Allocation (LDA), a probabilistic model, to extract latent semantic topics in the GPIs \cite{blei2003latent}. LDA model assumes that each document consists of several topics and that each topic is a distribution of words. 
Although every document in the GPI Corpus concerns privacy, there are several dimensions to this topic. Therefore, we partition GPIs into paragraphs to explore at a finer-grained level themes they contain.

Our GPIs are stored in text files, but there are no discernible patterns to extract the paragraph structure precisely for a text document. Therefore, we use two newline characters (\textbackslash{n}\textbackslash{n}) as a proxy indicator of a new paragraph unit to extract the paragraphs from a document. It results in paragraphs with a vast range (1-27,890 words) of length. To balance this range, we take a step further to divide the larger paragraphs into smaller paragraph units. We take a threshold of ten sentences and divide all the paragraphs we extracted in the previous step into the chunks of at most ten sentences. To filter out extremely small paragraphs, we remove all the paragraphs with less than nine words. With this technique, we are able to reduce the range to 9-1,133 words per paragraph. 
Each of these chunks forms a single input document unit for the LDA model. 

We apply the following steps to preprocess the input text segments: 
\begin{enumerate}
    \item We tokenize all the segments into uni-grams,
    \item We curate a custom list of stopwords. \textit{Stopwords} are words that carry very little information. For our context, words like ``article'', ``chapter'', ``number'' provide insufficient information and, we include typical stop words such as ``the'', ``is'' , ``an'' from gensim \cite{preprocessing} 
    We then remove all the stopwords from the text segments,
    \item We lemmatize all the tokens using WordNetLemmatizer \cite{wordnet},
    \item We remove all the tokens with less than three characters, and
    \item We filter out the tokens that occur less than 15 times and the ones present in more than 50\% of the documents.
\end{enumerate}

We generate a dictionary with the remaining tokens. We next compute the vector representation of each token using TF-IDF \cite{tfidf} and give it to the LDA model. One hyperparameter of the LDA is the number of topics (k) to be considered. We experiment with six values for k, equal to 5, 7, 10, 12, 15, and 20, and by manual analysis, we find that the cohesiveness of the resulting clusters decreases with an increase in the k. We also experiment with a combination of uni-gram and bi-gram inputs and find that uni-gram results in a higher coherence score. 


We manually interpret each output topic cluster by inspecting each topic's top ten relevant terms and the relevant documents. We get the best results for k equals ten and show our results in Table \ref{tab:topics_modeling}. We observe discernible meanings in eight clusters. Out of these eight clusters, four clusters show notable strong connections to the significant privacy concerns. These clusters cover Telecommunications, Bank and Finance, Healthcare, and Consumer Payments, which are intuitively common industries for privacy concern. 

We also observe subtle similarities between the Prosecution Process and Penalty clusters as the first describes the various aspects of prosecution, latter talks about the outcome of the prosecution. It is worth noting that these two topics suggest criminal laws. Given the broad inclusion criteria, we also include the criminal laws published by various jurisdictions that have sections devoted to privacy and data protection concerns.

\begin{table*}
\centering
\captionsetup{justification=centering}
  \caption{Prevalent topics across all the GPIs extracted using LDA.}
  \label{tab:topics_modeling}
  \begin{tabular}{ll}
    \toprule
    Topic & Example Keywords\\
    \midrule
    Telecommunication & Electronic, Operator, Network\\ 
    Penalty & Year, Imprisonment, Fine, Penalty\\ 
    Sources of authority & Court, Minister, President, Authority\\ 
    Debt \& Ownership  & Debt, Creditor, Property, Share\\ 
    Bank \& Finance & Institution, Bank, Investment, Financial\\ 
    Healthcare & Health, Medical, Care, Patient\\
    Consumer Payments & Consumer, Service, Credit, Electronic \\
    Prosecution Process & Prosecutor, Judge, Repeal, Order\\
    People and Regulations & Processor, Subject, Right, and Regulation\\ 
    National Level Authorities and Sanctions & Fine, Federal, National, and Sanction\\
  \bottomrule
\end{tabular}
\end{table*}



For the last two clusters, the top relevant terms point to a combination of topics instead of a single topic. The topic nine contains terms like ``processor'', ``subject'', ``person'', ``right'', and ``regulation'' which seems to be about People and Regulations. Similarly, with terms like ``fine'', ``federal'', ``national'', and ``sanction'', topic ten appears to be about National Level Authorities and Sanctions. 




\section{Discussion}


We discuss the issues in text analysis and legal enforceability of the GPI documents, and conclude with a discussion on the limitations of this work.


\textbf{Legal enforceability:} In the corpus, we mark each English translated document with whether the translation is completed by a human government translator, a 3rd-party government translator, or by our own machine translation. However, levels of legal enforceability of each type of document vary widely among countries and individual instances. For example, there are English translations of laws released by both third-party groups and government resources that proclaim that they are for informational use only and that the law is only legally enforceable in the original non-English language. In contrast, some jurisdictions appear to provide their laws in multiple languages but fail to specify which version of the document is legally enforceable.

The translations generated by the 3rd-party groups (e.g., private law firms) are less likely to be strictly legally enforceable than documents sourced directly from government websites. As the clarifications of legal validity or non-validity (or the absence thereof) vary wildly among the documents from all the different categories of sources, we are not able to definitively mark each document as technically legally enforceable in its current state. Thus, when we mark a particular document as ``in effect'', this is to say that the original law is ``in effect'', but this does not guarantee the legal accuracy and permissibility of all of the permutations of each document we provide within the corpus. These are only for informational and research purposes and cannot be assumed to be legally viable forms of these laws, regulations, and recommendations. This limits the applicability of this corpus as an up-to-date legal tool for use by legal counsel or by end-users seeking to guarantee their compliance with privacy laws, regulations, and recommendations.


\textbf{Limitations}: We acknowledge limitations of this work. First, this work takes a perspective that is centered on the English language, to match the expertise of the authors. This perspective, along with the sheer volume of text, required the use of machine translation for some of the document collection process and for the text analysis. Second, the lack of international standards for what constitutes a ``law'', ``regulation'', ``directive'' or other government document means that creating a truly exhaustive collection of GPIs is impractical. We mitigate that limitation through detailed collection rules and the use of online legal information resources, described earlier in the paper.


Regardless of these limitations, we provide first-of-their-kind observations and a corpus that others can build upon to study the international privacy landscape.

\section{Future Directions}

There are several opportunities for future work in this space. A collaborative effort of legal and multi-lingual experts could create a definitive ontology or an annotation system to arrange these documents into more meaningful categories for legal scholarship, thereby promoting further analysis of the metadata and text of the documents along with these categories. This corpus could be used to examine the legal interpretations and linguistic structures of similar legal sentences in different languages, yielding observations regarding the semantic and syntactic structure of these sentences and their legal intent. In addition, a large-scale human-reader analysis of the corpus (i.e., through crowdsourcing) presents an opportunity to draw observations in terms of similarities and differences among privacy standards around the world.

Given the recent success of NLP techniques in the legal and privacy domain, researchers can analyze the text documents of this corpus using various computational methods to uncover patterns in the structure and language of the documents in this corpus. Further, information retrieval techniques can be employed to develop a custom search engine and index all the GPIs from the corpus. Such a service could enable easy access to the relevant documents and leverage the metadata to create filters for jurisdiction type, year of enactment, and more. With advancements in deep learning-based question-answering, a chatbot or conversational agent can be developed to answer the privacy law-related questions of the users. 



\section{Conclusions}

In this paper, we introduce the Government Privacy Instructions (GPI) Corpus, a collection of 1,043 official GPIs from 182 jurisdictions around the world. We present our inclusion criteria for jurisdictions and GPIs. To the best of our knowledge, this is the first of its kind of study. We contribute text in both English and the original version of the documents. By leveraging text analysis tools, we present a large-scale empirical examination of the privacy laws and regulations published by governments to direct companies and organizations to pay attention to various aspects of consumers' privacy. We show how that attention has increased dramatically over time for some categories of Personally Identifiable Information (PII) more than others. We also see the signs of how this attention is distributed, resulting in only a few comprehensive privacy legislation. In addition, we observe that certain PII types appear together more often than others. Some correlations are intuitive (e.g., Biometric \& Genetic), while others (e.g., Race/Ethnicity \& Beliefs) are relatively unexpected. Overall, the results provide previously absent nuances for claims that privacy is receiving increased attention and regulation. 
Finally, by releasing the corpus, we provide a basis for further work to examine privacy regulation on a global scale.


\section{Acknowledgements}

We acknowledge Prof. Graham Greenleaf’s itemization of privacy laws as an inspiration for our work: Global Tables of Data Privacy Laws and Bills (7th Ed, January 2021) \citet{greentable}; Global Tables of Data Privacy Laws and Bills (6th Ed, January 2019) \citet{greenleaf2019global}. The present project began with collecting the set of laws that his tables refer to, continuing outward from that set. We are also grateful to Prof. Greenleaf for his correspondence while building this corpus. Additionally, the work reported herein was supported by the National Science Foundation under Grant \#CNS-1914444, \#CNS-1914486, and \#CNS-1914446.


\bibliographystyle{ACM-Reference-Format}
\bibliography{ref}




\end{document}